%% file: main.tex
\documentclass[letterpaper, conference]{IEEEtran}

\IEEEoverridecommandlockouts

\usepackage{geometry}
\usepackage{amsmath,amssymb,mathtools}

\usepackage{booktabs}
\usepackage{graphicx}
\usepackage{caption}
\usepackage{subcaption}
\usepackage{multirow}

\usepackage{siunitx}
\usepackage{microtype}

\usepackage{xurl}
\usepackage[hidelinks]{hyperref}
\usepackage[capitalise, noabbrev]{cleveref}

\input{macros}

\begin{document}

\title{Factorized Tactile Representation and Control for\\
  Sim-to-Real Manipulation}

\author{
\IEEEauthorblockN{
Siqi Shang\textsuperscript{1,2},
Bianca Aumann\textsuperscript{1},
Tye Brady\textsuperscript{1},
Joshua Migdal\textsuperscript{1},
Taşkın Padır\textsuperscript{1}
}

\thanks{${}^{1}$ Amazon Fulfillment Technologies \& Robotics,
Westborough, MA, USA.
\{biancjur, bradytye, jmigdal, ptaskin\}@amazon.com.}

\thanks{${}^{2}$ Department of Electrical and Computer Engineering,
The University of Texas at Austin, Austin, TX, USA.
siqi.shang@utexas.edu.}

\thanks{*This work was done when Siqi Shang was an intern at Amazon.}

\thanks{Taşkın Padır holds concurrent appointments as a Professor of Electrical
and Computer Engineering at Northeastern University and as an Amazon Scholar.
This paper describes work performed at Amazon and is not associated with
Northeastern University.}
}

\IEEEaftertitletext{%
  \vspace{-28pt}%
  \noindent
  \begin{minipage}{\textwidth}
    \centering

    \includegraphics[width=\linewidth]{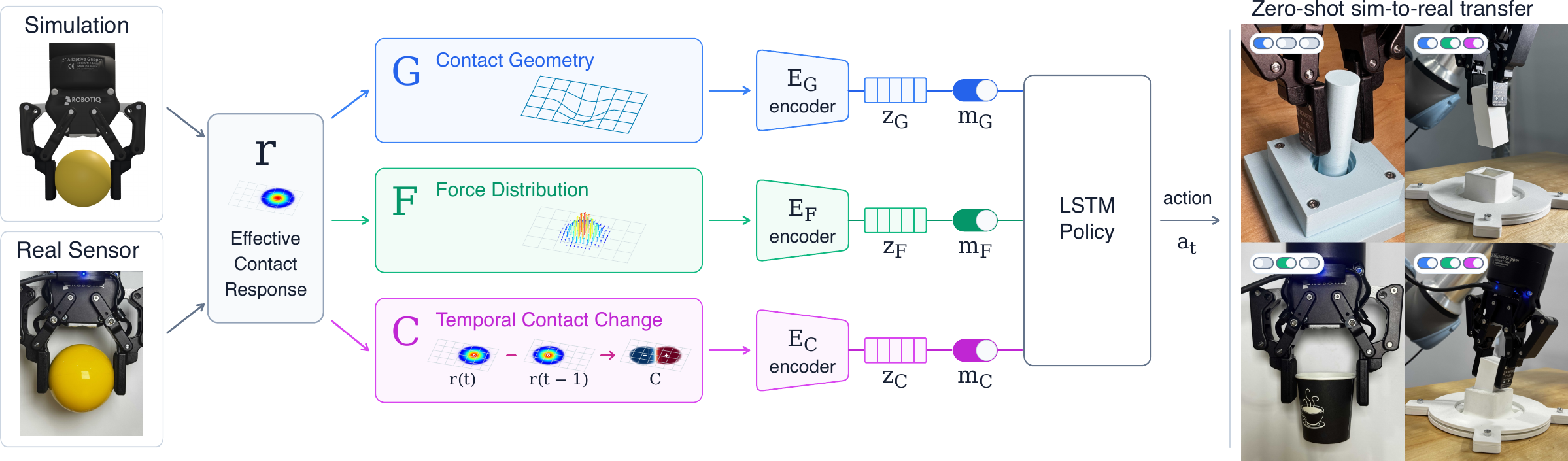}

    \par\footnotesize
    Fig.~1: Simulated contact and real tactile sensor readings are mapped to a
    common effective contact response \ecr{}, providing a shared tactile
    interface across the two domains. The response is factorized into contact
    geometry (\Grep), force distribution (\Frep), and temporal contact change
    (\Crep). To preserve this factorization through control, the Tactile Gated
    Policy encodes each representation separately and uses a tactile mask
    $\tmask$ to select which latents reach the LSTM policy. All eight mask
    configurations are trained within one policy per task, allowing the active
    tactile representations to be changed without retraining. The policies
    trained in simulation transfer zero-shot to real-world Tactile Alignment,
    Grasp-Force Regulation, and Adversarial Peg Insertion, revealing that
    different tactile representations benefit different interactions.
  \end{minipage}%
}

\maketitle

\setcounter{figure}{0}
\refstepcounter{figure}
\label{fig:overview}

\begin{abstract}
Tactile sim-to-real learning must bridge simulated contact and
device-specific sensor responses while preserving information needed
for control. We propose a factorized tactile representation and
control framework that maps normal force and contact patch to an
effective contact response recoverable from sensor readings. The
response is separated into contact geometry, force distribution, and
temporal contact change, with representation-specific encoding and
randomization. A Tactile Gated Policy preserves these representations
separately through control and operates over all mask configurations
without retraining. We evaluate the approach through response
reconstruction, spatial alignment, force regulation, and contact-rich
adversarial peg insertion in simulation and the real world, enabling
the utility and transfer reliability of different tactile
representations to be assessed independently. The approach achieves
$<$\SI{1}{\milli\meter} contact localization,
\SI{1.69}{\newton} force-tracking error on unseen geometries, and a
35\% improvement in real-world adversarial peg insertion over the
unfactorized response, with different tactile representations
benefiting different interactions.
\end{abstract}

\input{sections/intro}
\input{sections/related}
\input{sections/method}

\input{sections/policy}

\input{sections/experiments}

\input{sections/discussion}

\input{sections/acknowledgments}

\bibliographystyle{IEEEtran}
\bibliography{refs}

\end{document}

%% file: macros.tex
\newcommand{\Grep}{G}                 
\newcommand{\Frep}{F}                 
\newcommand{\Crep}{C}                 
\newcommand{\zG}{\mathbf{z}_{\Grep}}  
\newcommand{\zF}{\mathbf{z}_{\Frep}}
\newcommand{\zC}{\mathbf{z}_{\Crep}}

\newcommand{\encG}{E_{\Grep}}         
\newcommand{\encF}{E_{\Frep}}
\newcommand{\encC}{E_{\Crep}}
\newcommand{\tmask}{\mathbf{m}}       

\newcommand{\ecr}{r}                  
\newcommand{\taxcount}{c}             
\newcommand{\egain}{\beta}            
\newcommand{\pfield}{P}               
\newcommand{\spkern}{K}               
\newcommand{\gloc}{\eta}              
\newcommand{\gagg}{\gamma}            
\newcommand{\taxreg}{\mathcal{A}}     



%% file: sections/intro.tex
\section{Introduction}
\label{sec:intro}

Learning reliable contact-rich manipulation remains a central challenge
in robotics. Large-scale demonstration datasets have enabled substantial
progress, but acquiring diverse physical interactions requires
considerable hardware resources and human
effort~\cite{khazatsky2024droid}. Sim-to-real reinforcement learning
offers a scalable alternative through autonomous interaction in
simulation. For tactile manipulation, however, transferring a policy
requires transferring the contact feedback on which its behavior
depends. Physical tactile measurements reflect contact mechanics and
sensor-specific responses that are difficult to reproduce in
simulation. Enabling simulation-trained policies to use this feedback
reliably is therefore an important step toward scalable contact-rich
manipulation.

Recent advances such as TacSL and HydroShear demonstrate the potential
of tactile simulation for learning transferable contact-rich
behaviors~\cite{akinola2025tacsl,dang2026hydroshear}. An important remaining challenge is making the information available from compact capacitive arrays equally useful for sim-to-real control. These arrays provide
sparse measurements, yet their distributed responses can carry
fine-grained information about contact geometry, force distribution, and temporal
evolution. Simplifying touch into binary contact or compact
force-and-location summaries can facilitate transfer, but reduces the
information exposed to the
controller~\cite{su2024tactilerl,pan2026beyondbinary}. We instead seek
to retain distributed contact information in a common representation
across simulation and hardware, and to use that information in a
factorized manner throughout sensing and control. This distinction
matters because the information needed for a behavior and the
uncertainty it should tolerate are not the same for every component of
touch: geometric inference should tolerate variation in response
magnitude, force regulation requires loading magnitude to remain
informative, and relative temporal changes can remain useful despite
uncertainty in absolute sensor gain. Our objective is therefore not only to transfer a richer tactile response, but also to keep different forms of tactile information separately accessible to the policy rather than commit to one representation for every interaction.

We introduce factorized tactile representation and control for
capacitive tactile arrays (Fig.~\ref{fig:overview}). Normal force and
contact patch define an effective contact response \ecr{}, while
per-taxel calibration recovers the same response from sensor readings.
From this common response, we construct three tactile representations:
contact geometry (\Grep), force distribution (\Frep), and temporal
contact change (\Crep). To preserve this factorization through control,
we introduce the Tactile Gated Policy. It encodes the three
representations separately and supports representation-specific
randomization. A tactile mask keeps their latent features independently
selectable before they enter the policy. For each task, one policy is
trained across all eight mask configurations, allowing us to evaluate
each representation and change the active representations at deployment
without retraining. Using a parallel gripper
equipped with two 28-taxel capacitive pads, we validate the approach
from held-out response reconstruction to real manipulation. The
representation supports sub-millimeter contact localization with
$6.25\times5.71\,\mathrm{mm}$ taxel spacing, while the best force-aware mask achieves \(\SI{1.69}{\newton}\) real-world force-tracking MAE on unseen geometries. Tactile alignment,
grasp-force regulation, and adversarial peg insertion further
demonstrate how this common response supports precise and
contact-rich control through separately and jointly usable tactile
information.

Our contributions are:
\begin{enumerate}
    \item An effective contact response defined from normal force and
contact patch and recoverable from capacitive sensor readings,
providing a shared sim-to-real tactile interface while retaining
distributed spatial and loading information.

\item Factorized tactile representation and control that preserve
contact geometry, force distribution, and temporal contact change
through representation-specific encoding and randomization, while the
Tactile Gated Policy keeps them independently selectable within one unified
policy.

    \item A real-world evaluation on a parallel gripper with two sparse
capacitive pads, demonstrating response reconstruction on unseen contact
geometries, sub-mm contact localization, grasp-force regulation, and
challenging adversarial peg insertion.
\end{enumerate}

%% file: sections/related.tex
\section{Related Work}
\label{sec:related}

\begin{figure*}[!t]
  \centering
  \includegraphics[width=0.8\textwidth]{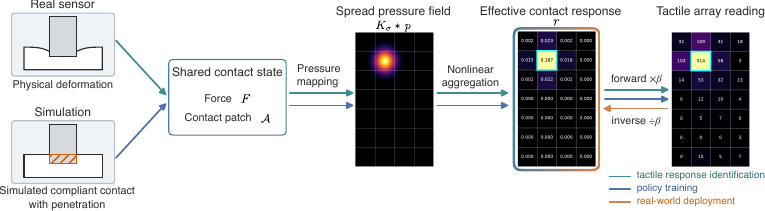}
  \caption{The response pipeline in both directions on a single calibration
    contact. From physical deformation on a real sensor or simulated compliant contact with penetration, force and contact patch produce a spread pressure field via an anisotropic Gaussian kernel, which is nonlinearly aggregated into the per-taxel effective contact response \ecr{}. The forward path multiplies by per-taxel gain $\egain$ to predict raw counts; the inverse path divides measured counts by $\egain$ to recover \ecr{}, separating the shared nonlinear contact response from unit-specific electronic gain.}
  \label{fig:pipeline}
\end{figure*}
\subsection{Tactile Simulation and Sim-to-Real Transfer}
\label{sec:rel-sim}

Tactile simulation has advanced rapidly for vision-based sensors,
where the measurement is an image that can be rendered directly.
TACTO~\cite{wang2022tacto} and Taxim~\cite{si2022taxim} model
GelSight-style sensors, while TacSL~\cite{akinola2025tacsl} and
Tac2Real~\cite{yan2026tac2real} bring visuotactile simulation to
GPU-parallel policy learning.

Taxel arrays require a different mapping from simulated contact to
distributed sensor responses. Kasolowsky and B{\"a}uml~\cite{kasolowsky2024}
model a soft tactile skin by reconstructing local penetration from
contact geometry and force, allowing contact to spread across multiple
taxels in a rigid-body simulator. A recent digital twin of the TSF-85
instead simulates a deformable sensing field and learns a mapping to
physical tactile counts~\cite{capacitivetwin2026}. Our approach avoids
reconstructing a deformation field. We define an effective contact
response from normal force and contact patch that is recoverable from
sensor readings and retains distributed information for policy
learning.

\subsection{Tactile Representations for Sim-to-Real Manipulation}
\label{sec:rel-rep}

Tactile representation is itself an important design choice for
sim-to-real manipulation. TactileRL shows that abstractions of
simulated tactile observations can transfer better than full tactile
images~\cite{su2024tactilerl}. Beyond Binary introduces a
physics-grounded center-of-pressure representation that retains force
and contact-location information while supporting zero-shot
transfer~\cite{pan2026beyondbinary}. These works show that successful
transfer need not require reproducing the full sensor observation.

A parallel line learns transferable tactile embeddings across sensors
and modalities, including AnyTouch~\cite{feng2025anytouch} and
TacX~\cite{higuera2025tacx}. Our objective is different. Rather than
compressing touch into one transferable representation, we preserve
contact geometry, force distribution, and temporal contact change
separately through encoding and control. This allows their utility
and sim-to-real behavior to be evaluated independently and their
availability to the policy to be changed without retraining.

%% file: sections/method.tex
\section{Factorized Tactile Interface}
\label{sec:representation}
Our tactile interface connects simulated contact to sensor readings through a common effective contact response. We first
define the contact quantities used to construct this response, then
factorize it into geometry, force, and temporal contact information.

\subsection{Shared Contact State}

\label{sec:shared}

We describe contact using two physically interpretable quantities:
total normal force $F$ and contact patch $\taxreg$. In simulation,
both are obtained from the contact solver. For real-world response identification,
they are measured from controlled contacts.
This abstraction is appropriate for the relatively stiff tactile pads
studied here. Gripper-position measurements indicate less than
\SI{1}{\milli\metre} of indentation at grasp forces up to
approximately \SI{60}{\newton}. We assume uniform pressure over the patch,
$\pfield=F/|\taxreg|$.

\subsection{Effective Contact Response}
\label{sec:response}

We map the pressure field to a per-taxel response through spatial
spreading and nonlinear aggregation. For taxel $j$ with region
$\taxreg_j$, we define the \emph{effective contact response} $\ecr_j$ as:
\begin{equation}
  \ecr_j =
  \left[
    \frac{1}{|\taxreg_j|}
    \int_{\taxreg_j}
    \bigl(\spkern_\sigma * \pfield\bigr)^{\gloc}
    \,\mathrm{d}\mathbf{x}
  \right]^{\gagg/\gloc}.
  \label{eq:response}
\end{equation}
The anisotropic Gaussian kernel $\spkern_\sigma$ models spatial
spreading across the pad. The exponent $\gloc$ controls the pointwise
nonlinearity, while $\gagg$ controls the overall force scaling.

The contact response $\ecr_j$ is a
phenomenological response variable rather than an estimate of pad
deformation, pressure, stress, or strain. Once $F$ and $\taxreg$ are
specified, the model does not use simulator-specific penetration
magnitude. The simulated and physical pads therefore do not need to
share the same deformation field.

Per-taxel sensor readings $\taxcount_j$ are related to the response by
$\taxcount_j=\egain_j\ecr_j+\varepsilon_j$, where per-taxel gain $\egain_j>0$
captures sensitivity and $\varepsilon_j$ denotes measurement
noise. We estimate the response as
$\hat{\ecr}_j=\taxcount_j/\egain_j$. Simulation therefore computes
\ecr{} from contact quantities, while sensor readings recover the same
response through the calibrated map, as illustrated in Fig.~\ref{fig:pipeline}.

\subsection{Response Identification and Operating Range}
\label{sec:identification}

We identify the response model from controlled real-world contacts
with known $F$ and $\taxreg$ and measured $\taxcount_j$. A single set
of shared parameters $(\sigma,\gloc,\gagg)$ is used for both pads and
all taxels. For each candidate $(\sigma,\gloc,\gagg)$,
\cref{eq:response} computes $\ecr_j$, and the optimal $\egain_j$ is
solved in closed form from $\taxcount_j$. We select
$(\sigma,\gloc,\gagg)$ by reconstruction error across both pads and
retain the corresponding $\egain_j$ for each taxel. The identified
values are $\sigma=(1.5,1.25)\,\si{\milli\meter}$, $\gloc=0.7$, and
$\gagg=0.5$.

The same shared response model is used for both pads throughout the
paper. Each pad's taxel-specific sensitivity is captured only through
$\egain_j$. \Cref{sec:exp-response} verifies that this shared model
reconstructs held-out contacts on both pads and generalizes to unseen
contact geometries. The calibrated mapping between $\ecr_j$ and $\taxcount_j$ remains
approximately linear up to \SI{20}{\newton} per taxel. Above this
range, $\taxcount_j$ compresses relative to the model prediction.
We therefore identify and evaluate the tactile interface within this
operating range.

\subsection{Tactile Representations}
\label{sec:representations}

We factorize \ecr{} into three tactile representations that expose
different contact information to the policy.

\paragraph{Contact geometry (\Grep)}
\Grep{} represents all 56 taxels as a point set with pad-local
coordinates and global pad pose. For pad $s\in\{L,R\}$, let
$\mathbf o_s\in\mathbb R^3$ denote its center and
$\hat{\mathbf u}_s,\hat{\mathbf v}_s\in\mathbb R^3$ its in-plane
unit axes in the global frame. For taxel $i$ on pad $s$, we compute
the local coordinates
$u_i=(\mathbf p_i-\mathbf o_s)^\top\hat{\mathbf u}_s$ and
$v_i=(\mathbf p_i-\mathbf o_s)^\top\hat{\mathbf v}_s$.

The normalized point coordinate is
$\mathbf q_i=[\bar u_i,\bar v_i,s_i]\in\mathbb R^3$, where
$s_i\in\{0,1\}$ identifies the pad. The point feature is
$\mathbf h_i=[\ecr_i,\mathbf o_s,\hat{\mathbf u}_s,
\hat{\mathbf v}_s]\in\mathbb R^{10}$.
We define
$\Grep=\{(\mathbf q_i,\mathbf h_i)\}_{i=1}^{56}$.
This 13-D representation per taxel preserves fine pad-local contact
location together with the global pose of each pad.

\paragraph{Force distribution (\Frep)}
\Frep{} retains the absolute response and its spatial distribution as
two $7\times4$ response grids, one for each pad.

\paragraph{Temporal contact change (\Crep)}
\Crep{} captures temporal change in contact through the per-taxel
normalized contrast of successive responses,
\begin{equation}
  \Crep_t =
  \frac{\ecr_t-\ecr_{t-1}}
       {\ecr_t+\ecr_{t-1}+\epsilon}.
  \label{eq:contrast}
\end{equation}
The contrast is bounded in $[-1,1]$. Responses below the noise floor
and the first observation of each episode are set to zero. Substituting $\taxcount_{j,t}=\egain_j\ecr_{j,t}$ gives
\begin{equation}
  \frac{\taxcount_{j,t}-\taxcount_{j,t-1}}
       {\taxcount_{j,t}+\taxcount_{j,t-1}}
  =
  \frac{\ecr_{j,t}-\ecr_{j,t-1}}
       {\ecr_{j,t}+\ecr_{j,t-1}},
  \label{eq:cancel}
\end{equation}
in the ideal noise-free case. Thus, \Crep{} is invariant to the
static per-taxel gain $\egain_j$.

\begin{figure*}[t]
  \centering
  \includegraphics[width=0.8\textwidth]{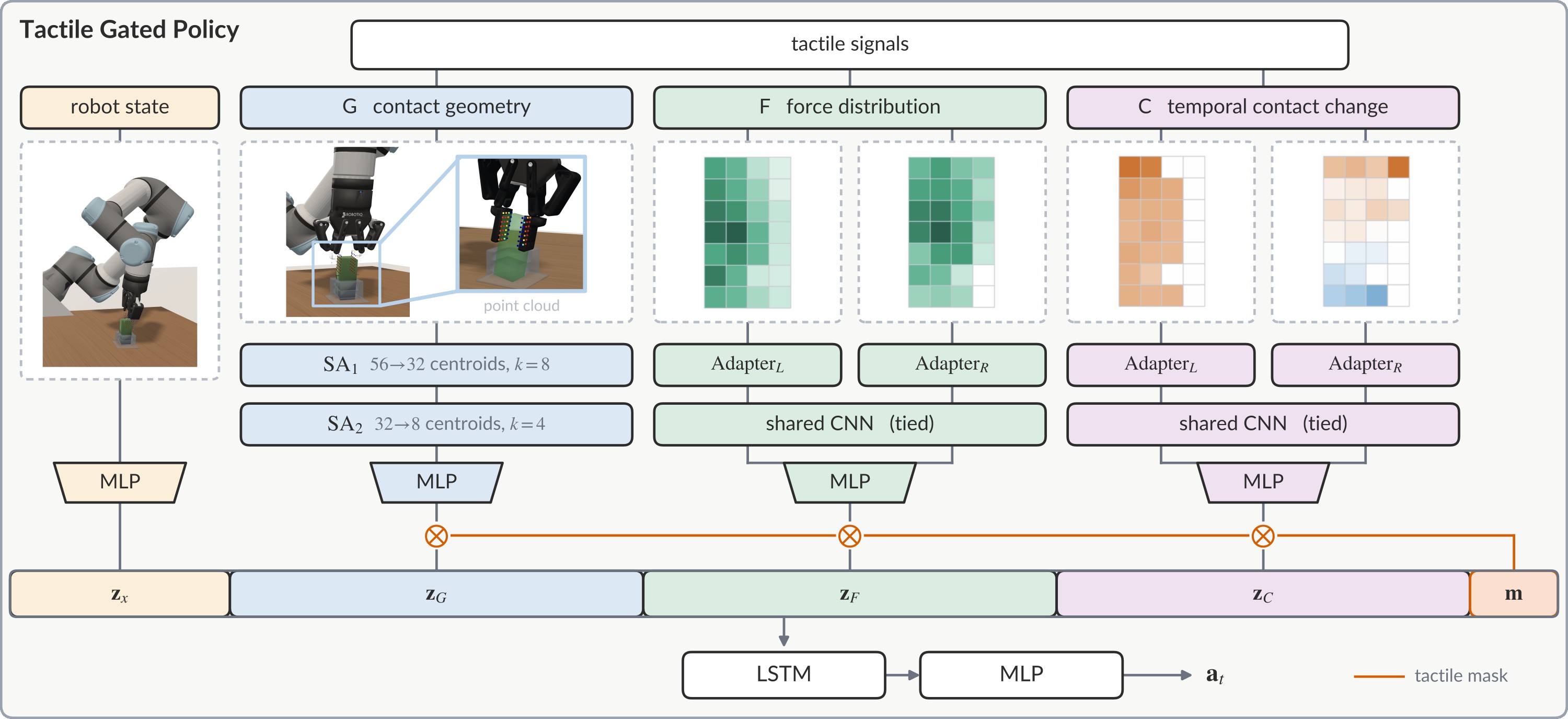}
  \caption{Tactile Gated Policy. Separate encoders produce geometry,
  force, and temporal-change latents. The mask $\tmask$ gates these
  latents before concatenation with the encoded robot state and
  mask itself as input to the LSTM actor.}
  \label{fig:architecture}
\end{figure*}

%% file: sections/policy.tex
\section{Tactile Gated Policy}
\label{sec:policy}

To preserve the tactile factorization through control, we introduce
the Tactile Gated Policy (Fig.~\ref{fig:architecture}). Each
representation is encoded separately and selected by a tactile mask
before entering the recurrent policy.

\subsection{Representation-Specific Encoders}
\label{sec:encoders}

All three representations are encoded into 64-D latents
(Fig.~\ref{fig:architecture}). The geometry encoder $\encG$ hierarchically processes the taxel point
set following PointNet++~\cite{qi2017pointnetplusplus}. Two
set-abstraction stages reduce the 56 taxels to 32 and then 8 points
using farthest-point sampling and local neighborhoods. Within each
neighborhood, learned features are summarized by both max pooling and
a response-weighted mean. A final global aggregation uses the same
pooling operations, followed by an MLP that produces
$\zG\in\mathbb R^{64}$. The force and change encoders, $\encF$ and $\encC$, operate on the
two per-pad grids. Each grid first passes through a pad-specific affine
adapter and then a CNN shared across the two pads. A fusion MLP
combines the two pad features into $\zF,\zC\in\mathbb R^{64}$.
$\encF$ and $\encC$ use the same architecture with separate weights.

\subsection{Tactile Mask}
\label{sec:gating}

A three-bit tactile mask
$\tmask=[m_{\Grep},m_{\Frep},m_{\Crep}]\in\{0,1\}^3$
selects which representation latents are available to the policy.
The selected latents are concatenated with the encoded robot state
and $\tmask$, then passed to an LSTM actor. The mask is fixed within each episode, and all eight configurations
are sampled during training. The all-zero mask uses proprioception
only. One policy is trained per task across all eight configurations,
allowing the contribution of each tactile representation to be
evaluated and the active representations to be changed at deployment
without retraining.

\subsection{Representation-Specific Randomization}
\label{sec:training}

Factorization also allows tactile uncertainty to be randomized
selectively during policy training. We perturb each representation
along dimensions that preserve its intended information.

For \Grep{}, the normal force is scaled independently on each pad by
$\lambda_s\sim\mathcal{U}(0.7,1.3)$ once per episode, while the
contact patch remains unchanged. This perturbs response magnitude
while preserving spatial contact structure.

For \Frep{}, the normal force is preserved while the contact patch is
rigidly misregistered in the pad plane. Each pad receives an
independent rotation and translation sampled once per episode. The
transformed patch is used to recompute the pressure field with the
original force. This perturbs the spatial distribution while
preserving absolute loading. Task-specific ranges are given in
\cref{sec:exp-controlled}.

\Crep{} receives no episode-level tactile randomization. Its normalized
temporal contrast is already invariant to the static per-taxel gain
$\egain_j$, as shown in \cref{eq:cancel}.

Sensor noise is applied upstream of all three representations at every
policy step. We sample independent count-space noise
$\varepsilon_{j,t}\sim\mathcal{N}(0,\sigma_c^2)$ with
$\sigma_c=\SI{4.2}{ct}$ and map it to response space through
$\egain_j$.

%% file: sections/experiments.tex
\section{Experiments}
\label{sec:experiments}

We evaluate three questions: \textbf{RQ1:} Can the effective contact
response bridge simulation and real tactile measurements?
\textbf{RQ2:} Does the response preserve contact geometry, force, and
temporal change? \textbf{RQ3:} How do the factorized representations
affect sim-to-real RL policy performance, separately and jointly?

\subsection{Experimental Setup}
\label{sec:setup}

\paragraph{Hardware and measurements}
Experiments use a UR5e arm with a Robotiq 2F-85 gripper and two TSF-85
capacitive tactile pads (Fig.~\ref{fig:rig}). Each pad has a $7\times4$ array over a measured $25\times40\,\mathrm{mm}$ active area with 
$6.25\times5.71\,\mathrm{mm}$ taxel spacing and noise of around $5$ reading counts. Readings are acquired at
\SI{500}{\hertz} and median-filtered over 21 samples.

Tactile data are collected with localized indentation and grasp-based
loading (Fig.~\ref{fig:rig}) \cite{shang2026forte}. For localized indentation
(Fig.~\ref{fig:rig}a), a hexagonal tip is pressed manually
along the pad normal while a ForceN force/torque sensor measures the
applied force with a reported error of \SI{0.2}{\newton}. For
grasp-based measurements (Fig.~\ref{fig:rig}b), the gripper compresses
an instrumented fixture with interchangeable contact geometries, referenced by a
\SI{20}{\kilogram}-range load cell 
at \SI{80}{\hertz}, with reported 
accuracy below $\SI{0.01}{\newton}$.

The rig base accepts interchangeable contact tiles. For response identification and evaluation (\cref{sec:exp-response}), six shapes are used
(Fig.~\ref{fig:rig}c): a full-face, an edge, two
circular indenters (diameters 4 and \SI{6}{\milli\metre}), and two
spherical indenters (radii 5 and \SI{20}{\milli\metre}). Localized
indentation and full-face loading identify the response
model; edge grasps test generalization to contact
family; flat-punches and spheres validate the simulated compliant-pad contact model. For grasp-force evaluation (\cref{sec:exp-controlled}), five different printed shapes replace the contact tiles (Fig.~\ref{fig:tasks}b).

\paragraph{Simulation and learning}
We use Isaac Sim 5.1.0 and Isaac Lab 2.3.2~\cite{mittal2023orbit}.
Policies are trained with recurrent PPO~\cite{huang2022cleanrl,schulman2017ppo}.
The actor uses an LSTM and only deployment-available observations.
The asymmetric, mask-conditioned critic uses privileged simulation
observations. Policies run at \SI{10}{\hertz}, with 12 physics steps
per policy step. The full tactile response stack accounts for
approximately \SI{9}{\percent} of simulation step time at 4096
environments per A100 GPU.

\paragraph{Policy evaluation}
Each experiment uses the mask convention defined in \cref{sec:gating}.
Within a checkpoint,
mask comparisons change only the available tactile information, not the
policy weights. 

\subsection{Effective Contact Response}
\label{sec:exp-response}

\paragraph{Contact datasets and evaluation protocol}

\begin{figure}[t]
  \centering
  \IfFileExists{figures/fig5_rig.pdf}{%
    \includegraphics[width=\columnwidth]{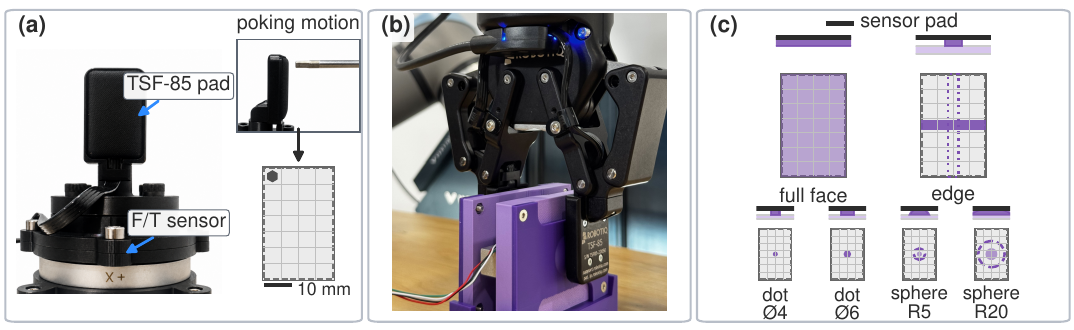}%
  }{%
    \IfFileExists{figures/fig5_rig.png}{%
      \includegraphics[width=\columnwidth]{figures/draft1/fig5_data_collection_setup.png}%
    }{%
      \fbox{\parbox[c][27mm][c]{0.94\columnwidth}{\centering
      \textbf{Contact data collection and pad-contact validation}\\[2mm]
      Poke rig with wrist F/T; edge and whole-pad grasping rig\\
      Load cell + HX711; flat punches and spheres}}
    }%
  }
  \caption{Contact data collection and pad-contact validation.
  Localized pokes use a wrist F/T reference; edge and whole-pad grasps
  use the load-cell rig. Separate grasping tests with flat punches
  (diameters 4 and \SI{6}{\milli\metre}) and spheres
  (radii 5 and \SI{20}{\milli\metre}) validate the simulated compliant
  pad-contact model. The simulated rig matches the physical geometry
  and loading; no direct physical contact-patch measurement is assumed.}
  \label{fig:rig}
\end{figure}

We collect synchronized tactile and force measurements from 280
localized pokes (five per taxel), 45 edge grasps, and 20 whole-pad
grasps (Fig.~\ref{fig:rig}c). Localized pokes use a
\SI{3.8}{\milli\metre} hexagonal indenter. Edge grasps use a
\SI{4}{\milli\metre} bar in horizontal and vertical
orientations stepped across the pad surface. Whole-pad loading is
recorded with the full-face contact, with 10 grasps per rig
orientation. Each contact is recorded for \SI{5}{\second}, yielding
approximately \SI{29}{\minute} of raw contact data. We evaluate two generalization settings. \emph{Spatial holdout} uses training pokes, selected edge lines, and whole-pad contacts, with
evaluation on unseen contact positions. \emph{Family holdout} excludes
all edge contacts from identification and model selection, providing a
stricter test on an unseen contact family.

\paragraph{Pad-contact model validation}
The grasping rig is reproduced in simulation with a compliant
pad-contact model with stiffness $k=\SI{5e4}{\newton\per\metre}$.
Validation covers 92 grasps on both pads with two flat punches and two
spheres, over \SIrange{1.5}{20}{\newton} per taxel. The simulated twin
achieves approximately 30-count RMSE with
median response-level deviation below \SI{3}{\percent}, and remains within \SI{10}{\percent} RMSE across a $20\times$ range of $k$, so precise stiffness identification is not required.

\paragraph{Response reconstruction}
\Cref{tab:baselines} separates held-out contact positions from an unseen
contact family. Under spatial holdout, learned encoders and $k$-NN
achieve lower poke error than the proposed model but higher edge
error. Removing all edges from fitting gives a stricter test: the
single response model shared across both pads achieves 31.4/30.8
counts on the left/right pads, compared with 92.4/73.8 for the MLP
and 114.8/85.7 for the CNN. The advantage is generalization to unseen
contact geometries, not minimizing interpolation error on training
pokes.

\begin{table}[t]
  \centering
  \caption{Median per-contact taxel RMSE in counts (left/right pad;
  lower is better). Spatial holdout fits on training pokes, selected
  edge lines, and whole-pad data. Family holdout excludes all edge
  data from fitting and model selection. Both edge evaluations use
  the same 20 held-out positions. Ablations use the spatial-holdout
  protocol. Methods within each block use matched fitting and
  evaluation splits.}
  \label{tab:baselines}
  \setlength{\tabcolsep}{3pt}
  \small
  \begin{tabular}{lcc}
    \toprule
    Model & Poke (L/R) & Edge (L/R) \\
    \midrule
    \multicolumn{3}{l}{\textbf{Baseline comparisons}} \\
    \addlinespace[2pt]
    \multicolumn{3}{l}{\emph{Spatial holdout}} \\
    Proposed \ecr{}     & 18.7 / 18.1 & 28.6 / 30.9 \\
    MLP encoder        & 13.3 / 13.9 & 83.5 / 63.6 \\
    CNN encoder        & 15.1 / 13.9 & 75.1 / 52.8 \\
    Kernel ridge       & 59.0 / 53.8 & 140.3 / 119.4 \\
    $k$-NN lookup      & 13.7 / 11.0 & 142.7 / 122.1 \\
    Depth proxy        & 49.1 / 49.0 & 84.2 / 74.3 \\
    \addlinespace[3pt]
    \multicolumn{3}{l}{\emph{Family holdout}} \\
    Proposed \ecr{}     & 20.4 / 19.1 & \textbf{31.4 / 30.8} \\
    MLP encoder        & 15.0 / 12.3 & 92.4 / 73.8 \\
    CNN encoder        & 17.3 / 11.5 & 114.8 / 85.7 \\
    \midrule
    \multicolumn{3}{l}{\textbf{Ablations: spatial holdout}} \\
    \addlinespace[2pt]
    Proposed \ecr{} (reference)
                       & 18.7 / 18.1 & 28.6 / 30.9 \\
    No spreading       & 22.4 / 25.2 & 43.9 / 50.3 \\
    Linear response    & 23.0 / 37.8 & 40.2 / 57.3 \\
    Hertz-dome profile & 18.2 / 19.0 & 32.5 / 33.8 \\
    \bottomrule
  \end{tabular}
\end{table}

The ablations test the main assumptions in the response model.
Removing spatial spreading or the nonlinear response substantially
increases edge reconstruction error. In contrast, replacing uniform
pressure with a Hertz-like pressure profile provides no improvement,
so we retain the simpler uniform-pressure assumption.

\subsection{Information Preserved by the Factorization}
\label{sec:exp-information}

\begin{figure}[t]
  \centering
  \includegraphics[width=\columnwidth]{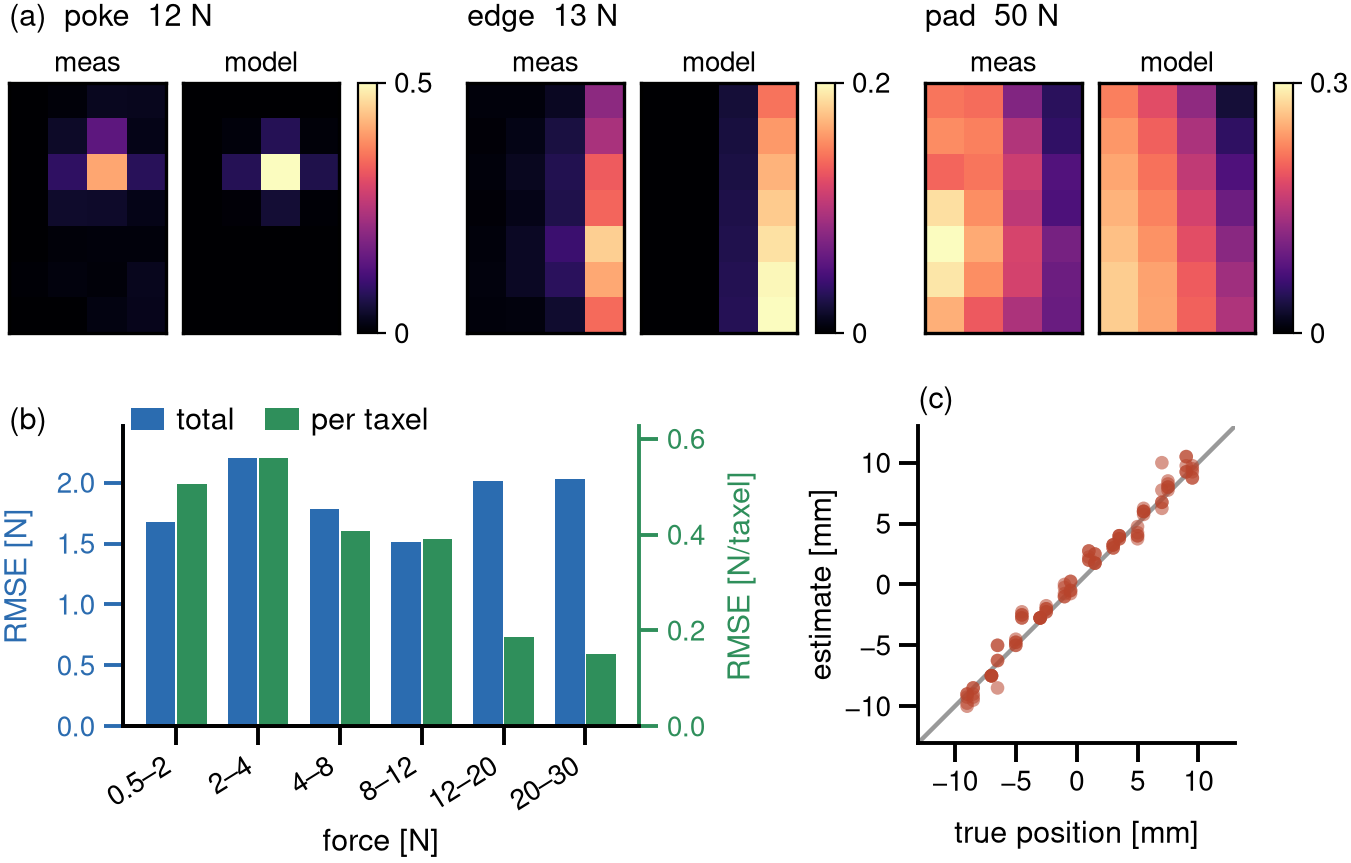}
  \caption{The effective response bridges physical tactile measurement while retaining
    task-relevant contact information. \textbf{(a)} measured and modeled count patterns for a 
    localized poke, edge, and distributed pad contacts. \textbf{(b)} held-out force error by true force bin, in total
    force and per activated taxel; error stays low inside the identified operating range and rises
    as the sensor approaches compression.
    \textbf{(c)} contact localization recovered from counts on a held-out fine scan, substantially finer
    than the taxel pitch.}
  \label{fig:information}
\end{figure}

We assess the spatial, force, and temporal information available
from the tactile response, independently of policy performance
(Fig.~\ref{fig:information}), using five-fold
cross-validation grouped by contact and stratified by source and pad.
Per-taxel calibration is fitted within each fold; whole-pad
data remain in training. These probes measure recoverable information,
not statistical independence between \Grep{}, \Frep{}, and \Crep{}.

\paragraph{Contact geometry}
A matched-filter probe localizes held-out contacts with $0.48\pm0.20\,\mathrm{mm}$ error (mean and
standard deviation of fold-wise median errors). A response-centroid
probe achieves $0.54\pm0.08\,\mathrm{mm}$. Both are below the
$6.25\times5.71\,\mathrm{mm}$ taxel spacing. These are response-level
localization tests; the alignment task below tests whether a learned controller
can exploit this spatial information.

\paragraph{Force distribution}
A recurrent force decoder estimates normal
force with \SIrange{1.7}{1.9}{\newton} RMSE over
\SIrange{0.5}{15}{\newton}, with variation across fold seeds.
Over the wider \SIrange{0.5}{80}{\newton} range, RMSE is
$3.03\pm0.57\,\mathrm{N}$ for the reported five-fold evaluation.
The wider-range result includes high-load response compression
beyond the declared operating regime (\cref{sec:identification}).
This is supervised force recovery from held-out physical contacts,
not a zero-shot transfer result.

\paragraph{Temporal contact change}
After the static gain cancellation in \cref{eq:cancel}, we evaluate
the remaining variation in \Crep{} over 144 constant-force plateaus.
The observed contrast standard deviation is 0.85--0.93 times that
predicted by the measured additive-noise model across a $100\times$
range of contact levels. This agreement shows that the residual
variation under static contact is well explained by sensor noise.

\subsection{Factorized Tactile Control}
\label{sec:exp-controlled}

We evaluate the factorized representations through three sim-to-real
RL tasks. Tactile Alignment isolates spatial information, Grasp-Force
Regulation isolates force information, and Adversarial Peg Insertion
tests their integration under changing contact constraints. All tasks
use the Tactile Gated Policy, with task-specific action spaces and
low-level robot control.

\begin{figure}[t]
  \centering
  \IfFileExists{figures/fig6_align_grasp_setup.pdf}{%
    \includegraphics[width=\linewidth]{figures/draft1/fig6_align_grasp_setup.pdf}%
  }{%
    \IfFileExists{figures/fig6_align_grasp_setup.jpg}{%
      \includegraphics[width=\linewidth]{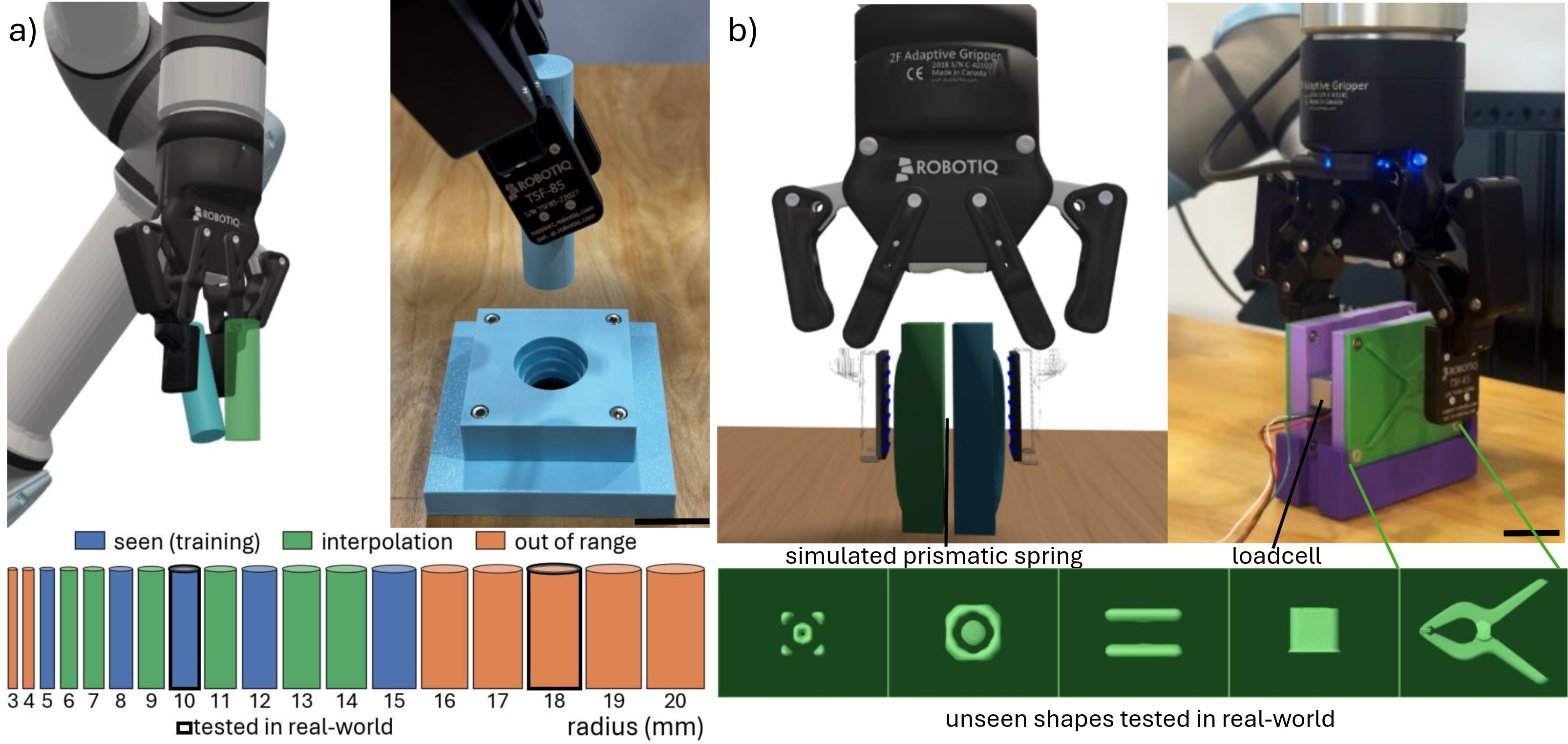}%
    }{%
      \fbox{\parbox[c][30mm][c]{0.97\linewidth}{\centering
      \textbf{Controlled manipulation setups}\\[2mm]
      (a) Tactile Alignment: grasp perturbations and stepped target\\
      (b) Grasp-Force Regulation: physical force rig and simulated
      two-plate prismatic spring}}
    }%
  }
  \caption{Controlled tactile manipulation.
  \textbf{(a) Tactile Alignment:} the policy aligns a cylinder with an
  unknown in-hand pose. Hardware evaluation records the deepest level
  reached in a fixture with 6, 4, 2, and \SI{1}{\milli\metre} radial
  clearance. \textbf{(b) Grasp-Force Regulation:} the policy commands
  gripper increments during a scripted hold-and-lift motion. The
  simulated object's two plates are coupled by a prismatic spring
  with randomized stiffness. An independent load cell supplies the
  hardware force reference. Scale bars: \SI{2}{\centi\metre}.}
  \label{fig:tasks}
  \vspace{-10pt}
\end{figure}

\subsubsection{Tactile Alignment}
\label{sec:exp-align}

The gripper holds an \SI{80}{\milli\metre} cylinder with an unknown
in-hand offset and tilt, and the policy commands Cartesian pose
increments to align its tip with a fixed target
(Fig.~\ref{fig:tasks}a). Training randomizes offsets within
$\pm\SI{5}{\milli\metre}$ and tilts within $\pm30^\circ$ across the
shown cylinder radii. For \Frep{}, the contact-patch registration is
additionally randomized independently on each pad by up to
$\pm5^\circ$ rotation and $\pm\SI{3}{\milli\metre}$ translation per
axis. The object pose is hidden from the actor and must be inferred
from tactile feedback. Simulation reports terminal position and orientation errors across the
radius sets. Real-world evaluation uses the
\SI{10}{\milli\metre} and unseen \SI{18}{\milli\metre} radii.
Alignment is evaluated by passage through the
\SI{1}{\milli\metre} and \SI{2}{\milli\metre} clearance levels.

\Cref{tab:align} shows that \Grep{} provides the spatial information
required for precise alignment. G and GC remain near or below
\SI{1}{\milli\metre} in simulation on unseen radii and achieve the
strongest hardware performance, with GC reaching the
\SI{1}{\milli\metre} clearance in 9/10 and 8/10 trials on $r10$ and
$r18$. \Frep{} provides only coarse alignment, while \Crep{} alone is
comparable to proprioception. The taxel-line reference performs
substantially worse, showing that the distributed tactile response
supports spatial estimation beyond a discrete taxel readout.

\begin{table}[t]
  \centering
  \caption{Tactile Alignment.
  Simulation reports terminal position/orientation error
  [mm/$^\circ$] on unseen interpolation and out-of-range radii
  (100 held-out resets per radius). Real-world reports successful
  \SI{1}{\milli\metre}- and \SI{2}{\milli\metre}-clearance trials
  out of 10 for $r10$/$r18$, with $r18$ unseen during training.
  \emph{Taxel-line} is an oracle-calibrated discrete taxel readout;
  its simulation entries are recovered in-hand pose RMSE rather than
  terminal policy error.}
  \label{tab:align}
  \footnotesize
  \setlength{\tabcolsep}{2.7pt}
  \renewcommand{\arraystretch}{1.03}
  \begin{tabular}{@{}lcccc@{}}
    \toprule
    & \multicolumn{2}{c}{Sim. [mm/$^\circ$]}
    & \multicolumn{2}{c}{Real [success/10]} \\
    \cmidrule(lr){2-3}\cmidrule(l){4-5}
    Input & Interp. & Out-of-range
          & 1 mm ($r10/r18$) & 2 mm ($r10/r18$) \\
    \midrule
    Prop. & 14.31/20.35 & 13.77/19.67 & 0/0 & 0/0 \\
    \textbf{G}
          & \textbf{0.62/0.65} & \textbf{1.06/1.49}
          & \textbf{9/7} & \textbf{10/10} \\
    F     & 1.83/1.97 & 2.50/2.74 & 3/3 & 9/8 \\
    C     & 9.91/14.25 & 9.77/14.04 & 0/0 & 0/0 \\
    GF    & 0.69/0.67 & 0.98/1.23 & 5/5 & 10/10 \\
    \textbf{GC}
          & \textbf{0.60/0.62} & \textbf{0.94/1.20}
          & \textbf{9/8} & \textbf{10/10} \\
    FC    & 1.69/1.80 & 2.15/2.31 & 2/3 & 8/9 \\
    GFC   & 0.74/0.65 & 1.15/1.40 & 9/2 & 10/8 \\
    \midrule
    Taxel-line & 2.20/3.97 & 2.21/4.46 & 1/0 & 2/1 \\
    \bottomrule
  \end{tabular}
\end{table}

\subsubsection{Grasp-Force Regulation}
\label{sec:exp-graspforce}

The policy regulates grasp force to a given target while the arm
executes a scripted hold-and-lift motion (Fig.~\ref{fig:tasks}b). It controls the Robotiq 2F-85 in position mode using its
integer $0$--$255$ position register, where one tick corresponds to
approximately $3.42$ mrad of finger motion. The action specifies an
incremental change to this position setpoint at each policy step.
Training uses 2048 environments and 404
object meshes from EGAD!~\cite{morrison2020egad} and
YCB~\cite{calli2015ycb}, with 50 validation and 50 test meshes held
out. To capture the varying compliance of real objects and make gripper
position an unreliable proxy for force, the simulated object consists
of two plates coupled by a prismatic spring whose stiffness is
randomized over a $25\times$ range (Fig.~\ref{fig:tasks}b). We additionally randomize gripper dynamics,
friction, object mass, and grasp pose. System identification measures
approximately \SI{100}{\milli\second} of gripper-command latency,
which is reproduced in simulation. Training targets span
\SIrange{2}{40}{\newton}.

For evaluation in simulation, we run 512 episodes per gate on the 50 unseen
test meshes and include tactile count noise corresponding to 
\SIrange{0.4}{0.6}{\newton} force uncertainty. In real-world, five unseen test shapes are printed and mounted to the
load-cell fixture (Fig.~\ref{fig:tasks}b), providing ground-truth contact force. Each shape is tested at targets of
1, 2.5, 5, 10, and \SI{15}{\newton}, with three grasps per shape,
target, and gate (600 trials total).

\begin{table}[t]
  \centering
  \caption{Grasp-Force Regulation.
  Simulation reports error on
  50 unseen test meshes; real-world evaluation reports pointwise settle MAE as
  mean $\pm$ SD across five printed test shapes.
  \emph{Reference} uses privileged true force in simulation and the
  best fixed gripper setpoint selected with load-cell hindsight on
  hardware. Bold highlights the consistently strong
  \Frep{}-containing gates.}
  \label{tab:graspforce}
  \footnotesize
  \setlength{\tabcolsep}{4pt}
  \renewcommand{\arraystretch}{1.03}
  \begin{tabular}{@{}lccc@{}}
    \toprule
    & \multicolumn{2}{c}{Sim. MAE [N]} & Real MAE [N] \\
    \cmidrule(lr){2-3}\cmidrule(l){4-4}
    Input & $\leq15$ N & 15--30 N & $\leq15$ N \\
    \midrule
    Prop. & 1.61 & 3.87 & $6.09 \pm 1.03$ \\
    G     & 1.66 & 2.26 & $3.80 \pm 1.02$ \\
    \textbf{F}
          & \textbf{0.95} & \textbf{1.33}
          & $\mathbf{1.91 \pm 0.47}$ \\
    C     & 5.48 & 6.13 & $11.78 \pm 13.25$ \\
    \textbf{GF}
          & \textbf{1.15} & \textbf{1.50}
          & $\mathbf{1.69 \pm 0.25}$ \\
    GC    & 1.45 & 2.72 & $4.45 \pm 1.04$ \\
    \textbf{FC}
          & \textbf{1.03} & \textbf{1.38}
          & $\mathbf{1.84 \pm 0.44}$ \\
    GFC   & 1.25 & 1.84 & $2.28 \pm 1.15$ \\
    \midrule
    Reference &0.29 & 0.37 & $1.32 \pm 0.33$ \\
    \bottomrule
  \end{tabular}
\end{table}

\Cref{tab:graspforce} shows that \Frep{} provides the information
needed for force regulation. F, GF, and FC achieve below
\SI{2}{\newton} mean real-world error and form the strongest group in
simulation, while removing \Frep{} increases hardware error to
$3.80$--$11.78\,\mathrm{N}$. GFC reaches
$2.28\pm1.15\,\mathrm{N}$, showing that using every tactile
representation is not optimal for this force-dominated task.
Simulation error remains nonzero even with privileged force feedback,
as the policy regulates force indirectly through gripper position
under randomized object dynamics and the identified
\SI{100}{\milli\second} actuation latency. Tactile policies
additionally receive sensor noise corresponding to approximately
\SIrange{0.4}{0.6}{\newton} force uncertainty.

\subsubsection{Adversarial Peg Insertion}
\label{sec:exp-insertion}
\label{sec:exp-detent} 

\begin{figure}[t]
  \centering
  \IfFileExists{figures/Fig7_Detent_Insertion_Setup.pdf}{%
    \includegraphics[width=\columnwidth]{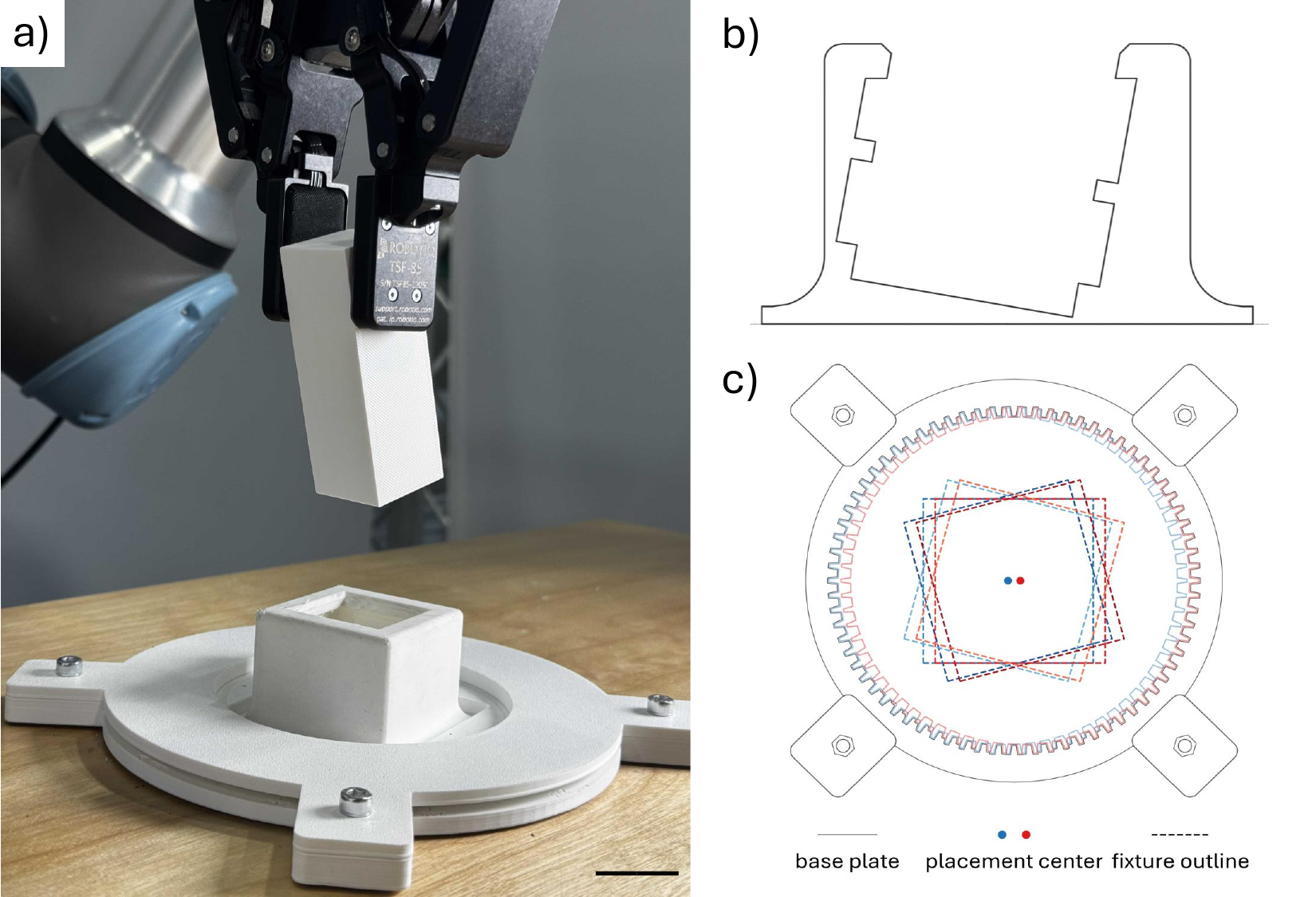}%
  }{%
    \IfFileExists{figures/draft1/fig_insertion_setup.png}{%
      \includegraphics[width=\columnwidth]{figures/draft1/fig_insertion_setup.png}%
    }{%
      \fbox{\parbox[c][30mm][c]{0.94\columnwidth}{\centering
      \textbf{Adversarial Peg Insertion setup}\\[2mm]
      Tilted bore and internal protrusions\\
      Randomized grasp and six fixture placements}}
    }%
  }
  \caption{Adversarial Peg Insertion.
  (a) The square peg must enter a tilted bore and pass two internal
  protrusions before becoming fully seated; grasp height, offset, and
  in-hand tilt vary between trials.
  (b) Cross-section of the tilted bore and internal protrusions.
  (c) The fixture is evaluated at two positions \SI{5}{\milli\metre}
  apart and three yaw angles ($-15^\circ$, $0^\circ$, $+15^\circ$),
  yielding six poses.}
  \label{fig:insertion_setup}
\end{figure}

The final task combines uncertain grasp and fixture poses with changing
contact constraints. A $30\times30\times80\,\mathrm{mm}$ square peg
enters a \SI{1}{\milli\metre}-clearance bore tilted
\SI{10}{\degree} from vertical and passes two internal protrusions before
becoming fully seated (Fig.~\ref{fig:insertion_setup}). Both grasp and
fixture pose vary and are hidden from the actor.

Training uses 8192 environments with compliant operational-space
control. Following reset-based exploration
methods~\cite{Khandate-RSS-23,omnireset}, we generate diverse states
along the insertion process to facilitate RL exploration. Fixture yaw
varies within $\pm15^\circ$, planar position within approximately
$\pm\SI{25}{\milli\metre}$, and grasp pose is also randomized. Arm
system identification follows~\cite{omnireset}. For real-world evaluation, the robot picks up the peg under
$\pm15^\circ$ rotational and $\pm\SI{2.5}{\milli\metre}$ positional
variation. We test six fixture poses: two positions
\SI{5}{\milli\metre} apart at $-15^\circ$, $0^\circ$, and
$+15^\circ$ yaw.
Success requires full seating. Simulation evaluates 1024 held-out
starts per condition.

We compare three alternative tactile representations.
\emph{Binary} retains only active-taxel locations. \emph{CoP},
adapted from~\cite{pan2026beyondbinary}, summarizes each pad by its
total response and response-weighted centroid. \emph{Response}
directly encodes the complete effective response $r$ without G/F/C
factorization. For the $+\mathrm{DR}$ baselines, all tactile domain
randomizations are applied jointly to the single representation; GFC
applies them separately to each representation.

\begin{table}[t]
\centering
\caption{Adversarial Peg Insertion.
Simulation reports success over 1024 held-out starts; hardware reports
30 trials per condition. \emph{Representation baselines} compare
proprioception only (Prop.; no tactile input), Binary Contact,
CoP~\cite{pan2026beyondbinary}, and the complete unfactorized Response
$r$ against GFC. CoP summarizes each pad by total response and its
response-weighted centroid; $r$ directly encodes the complete response
field. For CoP-DR and $r$-DR, all tactile domain randomizations are
applied jointly to the single representation, whereas GFC applies
them representation-specifically. \emph{Factorization ablation}
compares all non-empty proper subsets of G/F/C with GFC. Bold indicates
the highest result in each domain.}
\label{tab:insertion}
\footnotesize
\setlength{\tabcolsep}{2.2pt}
\renewcommand{\arraystretch}{1.03}

\begin{minipage}[t]{0.49\columnwidth}
\centering
\begin{tabular}{@{}lcc@{}}
\toprule
\multicolumn{3}{c}{\textbf{Representation Baselines}} \\
\cmidrule(lr){1-3}
Input & Sim. [\%] & Real [\%] \\
\midrule
Prop.  & 52.6 & 5/30 (16.7) \\
Binary & 59.9 & 9/30 (30.0) \\
CoP    & 64.3 & 2/30 (6.7) \\
CoP-DR & 42.7 & 1/30 (3.3) \\
$r$    & \textbf{70.1} & 14/30 (46.7) \\
$r$-DR & 65.7 & 14/30 (46.7) \\
\textbf{GFC} & 61.3 & \textbf{19/30 (63.3)} \\
\bottomrule
\end{tabular}
\end{minipage}
\hfill
\begin{minipage}[t]{0.49\columnwidth}
\centering
\begin{tabular}{@{}lcc@{}}
\toprule
\multicolumn{3}{c}{\textbf{Factorization Ablation}} \\
\cmidrule(lr){1-3}
Input & Sim. [\%] & Real [\%] \\
\midrule
G    & 53.4 & 11/30 (36.7) \\
F    & 60.8 & 14/30 (46.7) \\
C    & 57.2 & 8/30 (26.7) \\
GF   & 59.4 & 12/30 (40.0) \\
GC   & 60.4 & 14/30 (46.7) \\
FC   & 58.5 & 14/30 (46.7) \\
\textbf{GFC} & \textbf{61.3} & \textbf{19/30 (63.3)} \\
\bottomrule
\end{tabular}
\end{minipage}
\end{table}

\Cref{tab:insertion} shows a clear difference between simulation and
physical transfer. The unfactorized Response $r$ achieves the highest
simulation success (70.1\%), confirming that the effective response
contains sufficient information to learn the task. On hardware, however, GFC achieves the highest observed success at 19/30 (63.3\%),
compared with 14/30 (46.7\%) for Response $r$, 9/30 (30.0\%) for
Binary, and 5/30 (16.7\%) for proprioception alone. The representation favored in simulation is not the one that performs
best after transfer.

The factorization ablation shows a consistent pattern. Removing any
one representation from GFC reduces hardware success: GF reaches
12/30, while GC and FC each reach 14/30. \Crep{} is weak alone
(8/30), but adding it does not reduce success in any of the four matched comparisons
and improves three, consistent with temporal change providing
complementary information during transfer. Jointly applying all randomizations to an unfactorized representation does not
recover the GFC result: $r$ remains at 14/30 with DR, while CoP
changes from 2/30 to 1/30.

CoP reaches 64.3\% success in simulation but only 6.7\% on hardware,
below proprioception-only. Simulated collisions produce
abrupt lateral centroid shifts that are absent on hardware, suggesting
that CoP exposes contact transients that transfer poorly.
Proprioception-only still retains indirect contact cues through
OSC tracking deviations, which appears to be more
reliable in this setting. GFC instead preserves distributed spatial
and loading patterns across the pads, providing complementary cues
beyond a single magnitude and centroid. Its higher hardware success
(19/30) supports retaining richer contact information over a compact summary that may be sensitive to collision-response
mismatch.

Together, the three tasks show that tactile utility depends on the
interaction: \Grep{}-containing masks support precise alignment,
\Frep{}-containing masks support accurate force regulation, and the
complete representation achieves the highest success in integrated
insertion. Full tactile input is therefore not uniformly optimal across
tasks. Preserving the factorization through the policy keeps these
different forms of tactile information independently accessible
without retraining.

%% file: sections/discussion.tex
\section{Discussion}
\label{sec:discussion}

The effective response \ecr{} is a phenomenological quantity, not a measurement of the pad's deformation, pressure,
stress, strain, or mechanical state. Its
justification is operational: over the declared range it reconstructs held-out physical
measurements and preserves usable information, 
certified only below approximately \SI{20}{\newton} per taxel. 
The property we exploit is recoverability from the real sensor together with the separation of the three kinds of
information, not fidelity of appearance. A more faithful forward model could replace the
compact response without disturbing anything downstream, which consumes \ecr{} rather than the
model that produced it.

Because \ecr{} follows from a net force and a contact patch rather than a deformable pad state, it is
compatible with GPU-parallel policy training. More broadly, tactile information with different
transfer properties need not be collapsed into one policy representation: keeping it separately
addressable is what turns a representation choice into a deployment option.

%% file: sections/acknowledgments.tex
\section*{Acknowledgments}
The authors used Claude and ChatGPT to assist with coding and to improve the manuscript narrative, sentence structure, and vocabulary. All core algorithmic concepts, experimental findings, and final interpretations were verified and remain the sole responsibility of the authors.